\documentclass[conference]{IEEEtran}
\usepackage{cite}
\usepackage{amsmath,amssymb,amsfonts}
\usepackage{algorithmic}
\usepackage{graphicx}
\usepackage{textcomp}
\usepackage{xcolor}
\usepackage[ruled,linesnumbered]{algorithm2e}
\usepackage{float}
\usepackage{subcaption}
\usepackage{hyperref}
\usepackage{comment}

\begin{document}

\title{CommunityAI: Towards Community-based Federated Learning} 

\author{Ilir Murturi,
        Praveen Kumar Donta,
        and~Schahram Dustdar\\
{Distributed Systems Group, TU Wien}, Vienna, Austria\\
\{imurturi, pdonta, dustdar\}@dsg.tuwien.ac.at
        }

\maketitle

\begin{abstract}
Federated Learning (FL) has emerged as a promising paradigm to train machine learning models collaboratively while preserving data privacy. However, its widespread adoption faces several challenges, including scalability, heterogeneous data and devices, resource constraints, and security concerns. Despite its promise, FL has not been specifically adapted for community domains, primarily due to the wide-ranging differences in data types and context, devices and operational conditions, environmental factors, and stakeholders. In response to these challenges, we present a novel framework for Community-based Federated Learning called CommunityAI. CommunityAI enables participants to be organized into communities based on their shared interests, expertise, or data characteristics. Community participants collectively contribute to training and refining learning models while maintaining data and participant privacy within their respective groups. Within this paper, we discuss the conceptual architecture, system requirements, processes, and future challenges that must be solved. Finally, our goal within this paper is to present our vision regarding enabling a collaborative learning process within various communities. 


\end{abstract}


%
\IEEEpeerreviewmaketitle

\section{Introduction}

In recent years, Federated Learning (FL) has emerged as a promising paradigm in machine learning (ML), offering a unique solution to the problem of balancing data privacy and collaborative model training. FL enables ML models to be trained collaboratively across distributed devices while protecting sensitive data held by these devices. This approach has found applications in various domains, from healthcare to finance, aiming to harness the collective intelligence of decentralized data sources. However, the widespread adoption of FL faces multifaceted challenges such as encompassing scalability, data heterogeneity, security and privacy concerns, and resource constraints \cite{zhang2021survey}. 

Traditionally, training high-quality learning models requires well-labeled large datasets (i.e., data points are tagged or categorized). However, these datasets often contain sensitive information, making it impractical or insecure to share with centralized servers for machine learning purposes. FL, as first introduced by \textit{McMahan et al.} \cite{mcmahan2017communication}, addresses this challenge by providing a privacy-preserving approach for knowledge sharing among collaborative devices. The primary objective of FL is to enable knowledge transfer in the form of model parameters (e.g.,  the weights) between devices and without exposing the raw data. Each participating device trains a model locally using its data. Once locally trained, these models are uploaded to a central server which aggregates the model parameters (i.e., often by averaging them) and creates a global model with the knowledge of participating devices \cite{li2023federated}.

Members within communities can work together to tackle shared challenges \cite{hsu2022empowering}, and FL offers a technological solution that preserves data privacy and fosters collective intelligence. Community domains typically refer to specific areas or sectors within a community or society that share common interests, characteristics, or concerns. These domains can vary widely and may include fields such as wellness, health monitoring, education, healthcare, local government, nonprofit organizations, social services, and more. Essentially, community domains are the different aspects or sectors of community life where individuals and organizations work together to address specific needs and issues within that community. Nevertheless, a community is not limited only to social aspects; a community can be created even from a group of devices or sensors that aims to address specific issues (e.g., anomaly detection, fault classification, etc.). However, the large differences across community data sources are mostly the reason for the difficulty in implementing FL in such scenarios \cite{li2021survey}. These differences encompass variations in data types, contextual intricacies, device heterogeneity, operational conditions, environmental factors, and the varied interests of involved stakeholders. 

Despite its potential, FL has not been tailored to address the specific demands of community-based domains \cite{wen2023survey}. In a community context, FL assumes a high degree of data similarity across all FL tasks. This means that the data collected and utilized by different participants or devices in the community share common patterns, characteristics, or features that make them suitable for collaborative machine learning. For instance, let's assume the Fitness and Wellness Community where participants aim to weight loss and adopt healthier lifestyles. The community members use a variety of wearable devices such as smartwatches, fitness trackers, and health monitors. These devices collect data on metrics like heart rate, steps taken, sleep quality, and more.  Each device may produce different data structures. Moreover, devices even from the same manufacturer and version may produce various results due to the heterogeneous environmental and operation conditions. In some cases, the data collected or available for training may not be very similar in terms of content, characteristics, or patterns. As a result, such potential data variations can lead to negative knowledge transfer which may affect model performance. Therefore, there is a need for adapted approaches for FL that accommodate data heterogeneity by allowing specialized submodels for groups of devices that share similar data structures or characteristics. 

In the literature, few works attempted to address the above-discussed challenges. For example, \textit{Hiessl et. al.} \cite{hiessl2020industrial,hiessl2022cohort} introduced an industrial FL, where knowledge exchange can be performed based on data similarities in manufacturing industry data. This approach enables clients to select ML models according to their preferences for FL for industrial time series data. Their approach can recognize identical data, and distribute them accordingly. However, a few characteristics can be used to identify identical data. Nevertheless, this method does not specify the community identification process. In addition, the accuracy of identifying similar communities has not been verified. Moreover, this work is designed for industrial applications and has not been verified for other applications such as healthcare and smart cities. 

In response to these challenges, we introduce a novel framework called CommunityAI that seeks to organize participants into communities based on shared interests, expertise, data similarities, or characteristics. These communities collectively contribute to the training and refinement of ML models while ensuring the utmost protection of data and participant privacy within their respective groups. We assume that stakeholders have the ability to establish diverse FL communities, where they provide their respective ML models, tasks, and metadata. FL communities are then created based on similar configurations (i.e., such as device types, FL algorithms, ML models, and objectives). In order to address potential data dissimilarities and prevent negative knowledge transfer caused by model updates, we use the concept of FL cohorts (i.e., FL community subsets). This approach allows knowledge sharing exclusively within these cohorts and among models that relate to a community, where multiple characteristics are considered. These characteristics not only identify data similarity, but also consider multiple characteristics like shared interests, expertise, data similarity or characteristics, device location, or application. The community detection can be done autonomously and on-the-fly, which further enhances model learning speed. In this context, the major contributions of this paper are summarized as follows:

\begin{itemize}
\item We introduce a conceptual architecture and explore system requirements and underlying processes. The proposed architecture is designed by considering the distributed nature of three-tier infrastructures \cite{pujol2023edge, casamayor2023fundamental}.

\item {We outline various applications across different domains that can leverage the advantages offered by the CommunityAI framework.}

\item We present potential research directions that can foster novel studies in this field and overcome the current limitations.
\end{itemize}

The remaining sections are structured as follows. Section \ref{sec:background} gives a brief overview of community domains and FL, data source heterogeneity, and possible applications that may benefit from the CommunityAI framework. Section \ref{sec:Requirements_arch}  gives an overview of the CommunityAI framework, system requirements, and the architecture and processes of the proposed framework. In Section \ref{sec:futurechallenges}, we outline research challenges and future directions. Section \ref{sec:conlusion} concludes the paper.







\section{Background and Motivation}
\label{sec:background}
This section gives an overview of \texttt{(i)} community domains and FL, \texttt{(ii)} data sources, and \texttt{(iii)} CommunityAI applications.

\subsection{Community Domains and FL}
\begin{figure}[h]
    \centering
    \includegraphics[width=\columnwidth]{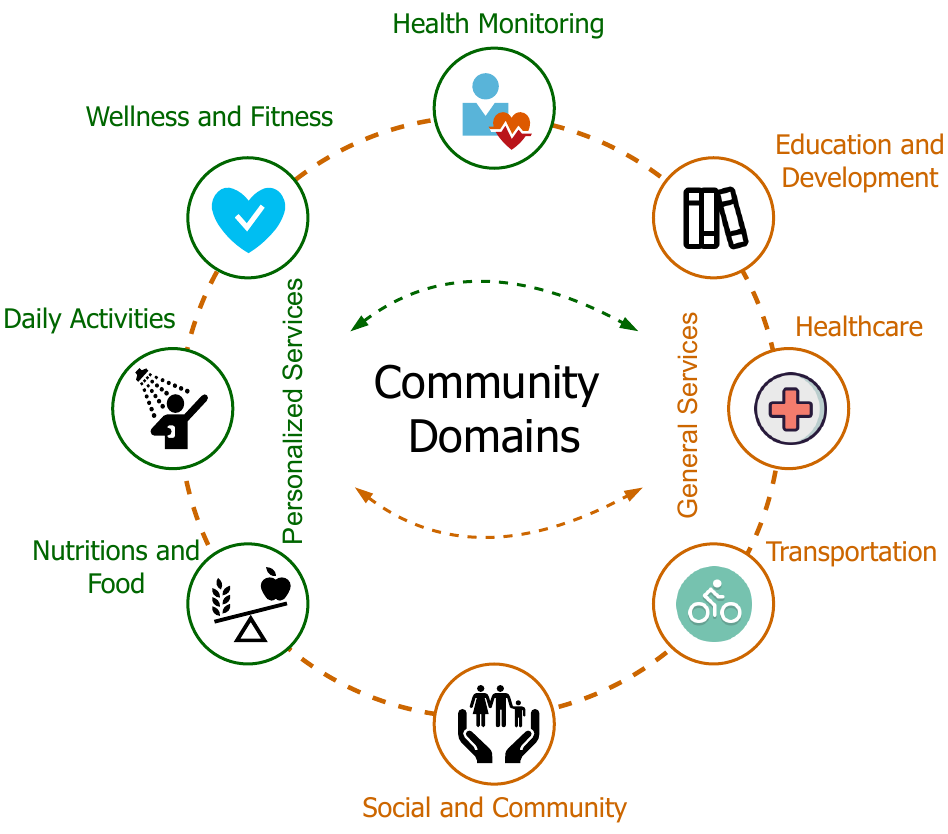}
    \caption{Community domains. }
    \label{fig:domains}
\end{figure}

In Figure \ref{fig:domains}, we name a few possible domains that can be formed and cater to community needs. Several software services can be developed for each such domain, which can be categorized as \textit{personalized} or \textit{general services}. For instance, the Wellness and Fitness Community domain may encompass physical, mental, or emotional health aspects within a community. This could involve (sub-)communities such as fitness, sports activities, stress management, and more. Essentially, such a community gathers virtually individuals with a common interest in promoting a healthy and active lifestyle. It brings together fitness trainers, wearable device users, nutritionists, and health enthusiasts who aim to harness the power of technology and data to enhance their well-being. Users can access personalized guidance, collaborate with trainers, and benefit from data-driven insights to improve their health. However, when considering FL approaches, the difficulty arises from various data sources and data structure heterogeneity. FL typically involves multiple distributed data contributors, such as humans, wearable devices, sensors, and different fitness trackers. 

The traditional FL approaches can be challenging because they may not adequately capture nuanced variations and individual users' preferences. For example, one user's heart rate data from a high-end smartwatch may differ significantly from another user's data collected from a basic fitness tracker. Consequently, a single model might struggle to provide accurate and personalized recommendations for both users. By allowing specialized submodels (i.e., cohorts) for groups of devices that share similar data structures or characteristics, users may receive more accurate and relevant recommendations for their devices and data sources. Furthermore, these specialized submodels reduce the computational load on each device, as they only need to communicate with their respective submodel rather than the global model. As a result, the system should scale more effectively to accommodate more communities, more users, and data sources.  


\subsection{Data sources}\label{subsec:datasources}
Embedded computing devices (ECD) are mechanical or electronic devices programmed to perform a specific task \cite{leveson1991software}, and they are the primary source for data in computing continuum systems. In our daily lives, we use ECDs such as microwaves, embedded washing machines, engineering calculators, digital cameras, digital door locks, health care devices, vehicle components, etc. These devices are designed with multiple hardware components including sensors and actuators, communication modules, power supply, tiny memory, and processors (either microcontroller (MC) or microprocessor) \cite{huang2017energy}. Due to high Internet availability, ECDs enable Internet connection further increasing remote accessibility. The Internet also helps to store a large amount of device activity data in the cloud. This further increases predictive maintenance, working conditions such as performance and downtimes. Further, depending on their hardware availability and range of features, they are divided into small-scale (using 8-bit MCs), medium-scale (using 16-bit or 32-bit MCs, or multiple 8-bit MCs), or sophisticated-scale (with complex software codes and hardware components). These enhancements and complex hardware and software features have attracted attention from many fields in recent days, and wearable devices are rapidly gaining popularity. 

\begin{figure}[h]
    \centering
    \includegraphics[width=1\columnwidth]{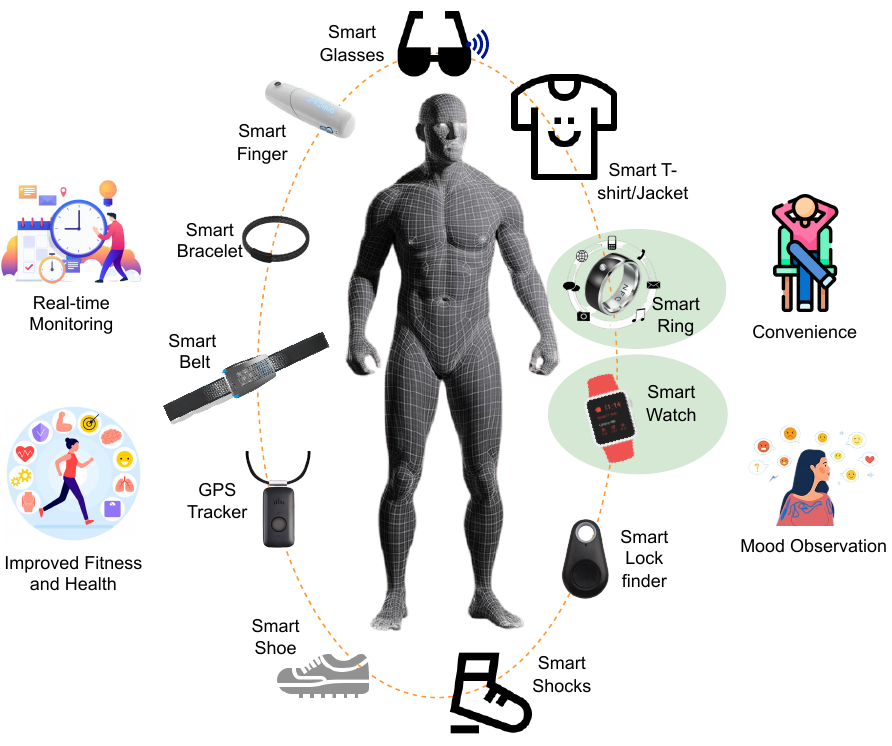}
    \caption{An overview of wearable ECD and their benefits.}
    \label{fig:Wearable}
\end{figure}

For example, wearable devices are becoming more common these days, as they are part of our daily lives. They can capture data by interacting with users and other devices through equipped sensors, processors, and connectivity features ranging from wearable devices such as smartwatches, fitness trackers, smart shoes, augmented reality glasses, and many more as shown in Figure~\ref{fig:Wearable} \cite{rodrigues2018enabling}. Figure~\ref{fig:Wearable} also shows various wearable devices, their data similarities (i.e., highlighted with green color) and differences, along with their benefits. Wearable devices monitor several metrics, including physical activity, regular health metrics (such as heart rate, sleep, blood pressure, and stress levels), and provide quick access to information through notifications. Their portability and constant presence make them ideal for bridging users with a wide network of devices. In addition to convenience, these devices can monitor fitness and health, track mood swings, communicate with people and other devices remotely, provide real-time analytics, and ensure safety monitoring \cite{nahavandi2022application}. Recent statistics (by GlobalData\footnote{\url{https://www.medicaldevice-network.com/comment/wearable-technology-iot}, Last accessed: \today}) show that the wearable devices market is generated \$59 billion in 2020. These studies also confirm that this growth will continue to more than \$156 billion by 2024 with a 24.6\% growth rate. The limitations of these devices in terms of memory and computation further make them dependent on other devices such as the edge, mobile phones, or cloud centers.

\subsection{CommunityAI Applications}
The CommunityAI does not limit itself to specific applications, and it supports a large number of applications, whereas only a few are discussed here. 
\subsubsection{Wellness and Fitness Application} Staying fit and healthy has become a top priority for many individuals. A cutting-edge system has emerged where people get daily recommendations for their health and fitness journeys. End-users, via their dedicated apps installed on their smartphones, may join various communities provided by the CommunityAI (e.g., Wellness and Fitness Community), actively contribute their data to this collaborative effort, and get personalized sport and activity suggestions based on data collected from various wearable devices. These devices allow users to provide context information and many metrics, including heart rate, steps taken, sleep quality, and more data that might be provided from other nearby wearable devices (as illustrated in Fig.~\ref{fig:Wearable}). 

The CommunityAI aggregates and analyzes data, transforming it into actionable insights. By employing learning algorithms, it builds individualized user profiles that consider factors such as fitness levels, health objectives, and even daily schedules. This comprehensive understanding enables the system to offer recommendations that are not only effective but also adaptive, taking into account a user's ever-changing fitness journey. For instance, a high-intensity enthusiast might receive guidance on rigorous High-intensity interval training (HIIT) workouts. At the same time, someone focused on weight loss might be directed towards a combination of calorie-conscious dietary plans and cardio exercises. 

\subsubsection{Governing Computing Continuum Systems}
Computing continuum systems are more heterogeneous since they include a variety of devices. Besides computing business data, these devices also generate a huge amount of data through logs. This data is analyzed to monitor their conditions, so that they can be used more efficiently while avoiding downtime. However, these systems generate a greater variety of data. For example, this system consists of custom logs for devices, network information, infrastructure details and location information \cite{donta2023governance}. In this case, each of these categories can be considered a distinct community. Further, each of these categories can be subcommunities because of their heterogeneity. For example, a computing continuum system contains a variety of devices including EDCs, sensor nodes, IoT, Edge/Fog nodes and cloud servers. It is important to recognize that each device has its own characteristics and capabilities. The same applies to network infrastructure. Based on the requirements, it can use LoRa, Bluetooth, WiFi, 4G, 5G, or 6G communication mediums. In this scenario, it is not feasible to use the same learning approach. Also, it is not efficient to generate a model by combining all these data. In this case, CommunityAI can offer more advantages, such as an efficient way of generating the model and analyzing the data. 

\subsubsection{Industrial Automation}
Industrial automation is the application of technology and control systems to simplify manufacturing with minimal human intervention while enhancing efficiency and productivity. Because of the coexistence of different equipment, machinery, and systems, this application was heterogeneous. This diversity can be challenging to manage since there are so many communication protocols, data formats, and interfaces to deal with. Using all these diverse data to train the model may result in an inefficient learning process. The accuracy of models, however, is more likely to lead to the highest productivity, which further enhances profits in the industry. It is not possible to get high-accuracy models from existing learning models because they are not efficient at fulfilling the current demands of industrial automation. Nevertheless, \textit{Hiessl et. al.} \cite{hiessl2020industrial} designed a model entitled Industrial federated learning to address this challenge, but this model needs many predetermined setups. One such requirement is to define FL cohorts and populations, but there are no specific methods for doing so. This model still needs to be enhanced to be more adaptive and dynamic in real-time. Nevertheless, CommunityAI can define these partitions autonomously in run-time based on the type of information. For example, the FL cohort in CommunityAI can be performed according to Data Distribution Service (DDS) principles \cite{donta2023towards}. 

\subsubsection{Healthcare}

Healthcare is another interesting field in which diverse data can be seen. The structures, formats, and sources of a wide range of healthcare data types-ranging from electronic health records (EHRs) to medical imaging and genomic data strikingly varied in this domain. While EHRs contain structured data fields, unstructured physician notes, and diagnostic images, genomic data contains DNA sequences and genetic variations. Moreover, data originates from a wide variety of healthcare facilities, such as hospitals, clinics, laboratories, and wearables, each adhering to unique standards and storage methods. Healthcare faces formidable challenges due to heterogeneous data. Unlike previously discussed applications, healthcare data is processed separately. For example, there are a variety of models to evaluate only physician notes, and various algorithms for DNA sequences. There is no common platform in the healthcare industry where all patient data can be analyzed. So, patient reports may not be received on time due to multiple platforms and the fact that they are not all available at one location. This opens the doors for all a patient's or multiple patients' data to be analyzed using a common platform with parallel computing nodes. This further simplifies the analyzing process for different metrics to decide the root cause of a patient's disease. Considering this scenario, the CommunityAI model would be more appropriate, since it can automatically determine where the data will be computed and produce accurate models based on the data. 

\section{CommunityAI: Requirements, Architecture, and Process}
\label{sec:Requirements_arch}
This section gives an overview of the CommunityAI framework. We discuss \texttt{(i)} system requirements (i.e., notations, stakeholders, and metadata), \texttt{(ii)} FL within CommunityAI, and \texttt{(iii)} the architecture and processes of the proposed framework.

\subsection{Notations, Stakeholders, and Metadata}
The CommunityAI foundational terminology is built upon the FL notation originally proposed by \textit{Bonawitz et al.} \cite{bonawitz2019towards}. This notation encompasses critical elements, such as devices, FL servers, FL tasks, FL populations, and FL plans. Devices represent various hardware platforms, including edge devices and mobile phones, equipped with FL clients that carry out the computational tasks required for training and assessing learning models. An FL client establishes communication with the CommunityAI to execute FL tasks associated with a specific FL population (i.e., we refer also as \textit{FL Community}). FL population refers to a globally unique identifier representing a shared learning objective across multiple FL tasks. The CommunityAI aggregates outcomes, which are model updates, stores the global model and then distributes it to FL clients within the designated FL population.
Furthermore, an FL plan is associated with an FL task and serves as the instructions for its federated execution. Such a plan guides the CommunityAI and the participating FL clients on federated execution. Later, we consider FL cohorts (i.e., as in the proposed approach \cite{hiessl2020industrial}) to enable that group multiple FL tasks within the same FL population and with similarities in data structure (described in Section~\ref{fl_communities}).

Within the CommunityAI framework, several stakeholders are involved in creating and managing communities. These stakeholders include:
\begin{itemize}
\item{\textit{Users or Participants}}: Users are the core stakeholders who join and actively participate in communities based on their interests, expertise, or data characteristics. They contribute data, engage in collaborative model training, and benefit from the insights and recommendations generated within their respective communities. Furthermore, participants can include domain experts, data scientists, fitness trainers, healthcare professionals, or individuals with specialized knowledge. 

\item{\textit{Sensory Data Contributors}}: Devices and sensors are active stakeholders as sensory data contributors. They generate valuable data shared within the community for collaborative ML. These devices play a critical role in providing the raw data that forms the foundation of insights and recommendations. More details about data sources are discussed in Subsection~\ref{subsec:datasources}.

\item{\textit{Community Creators}}: Community creators are individuals or organizations responsible for initiating and establishing communities within the CommunityAI framework. They define each community's purpose, objectives, ML models, tasks, and guidelines and often lead in guiding community activities.

\end{itemize}

Several requirements exist in the CommunityAI framework that must be fulfilled before fully operationalization. In order to facilitate collaboration among FL clients, we recognize the need to disseminate metadata that describes all involved stakeholders within the CommunityAI framework. First, we assume that manufacturers provide metadata that describes sensors and devices. On the other side, human contributors can create and set their personal data information on their own profiles. Second, community creators and FL clients may have specific criteria for collaborating with other clients. This could include requirements such as data quality, expertise level, trustworthiness, or alignment with particular objectives. Therefore, without proper collaboration criteria, the FL system may not effectively filter and select appropriate participants, leading to inefficient or potentially harmful collaborations. 

\subsection{FL Communities}
\label{fl_communities}
FL client choice is a critical aspect in the FL process, which shortens processes like training and evaluation \cite{nishio2019client}. We consider that stakeholders can form various FL communities. This means that a stakeholder for an FL community provides ML models, tasks, metadata, etc. The next step enables creating and assigning submitted tasks to their respective populations. Essentially, an FL population is built with tasks with the same configurations (e.g., device type, FL algorithm, ML model, objectives, etc.). When the configuration of a new task matches the existing population, it is included within the same population. Otherwise, a new population is created. It is essential to take into account a valid FL setup, where identical algorithms and models must be applied to the common data format. We do consider as a potential solution the proposed approach in \cite{hiessl2022cohort}, where a population is split into cohorts representing clusters of tasks with similar data distributions. In this way, within an FL community, FL clients exclusively exchange updates with a subset of FL clients whose submitted FL tasks are in alignment with the same FL communities and cohort. Note that the proposed approach \cite{hiessl2022cohort} considers only an industrial domain and data similarity aspects. In our approach, we extend FL community selection based on multiple metrics instead of data similarity, such as shared interests, expertise, and data characteristics. For the data similarity, a potential solution can be topic modeling \cite{weston2023selecting} for achieving highly accurate community separation.

\begin{figure}[h]
    \centering
    \includegraphics[width=\columnwidth]{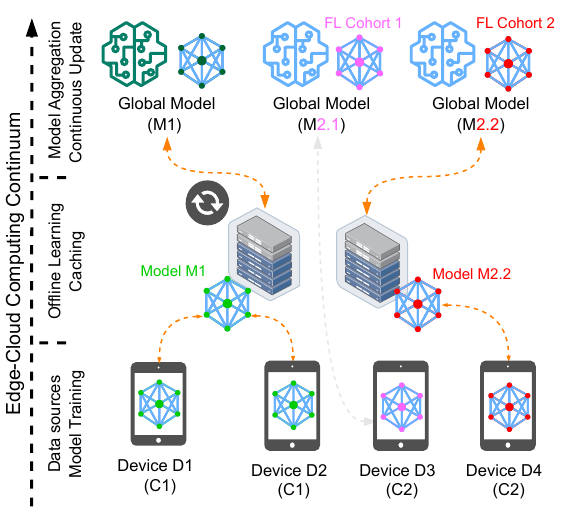}
    \caption{FL within CommunityAI.}
    \label{fig:fl_com}
\end{figure}

From a practical point of view, consider a group of devices that work together to solve a specific learning problem. This shared learning objective is typically related to a particular task, such as anomaly detection, natural language processing, recommendation systems, improving accuracy and device performance, etc. For example, consider a smartwatch or bracelet where the manufacturer may want to improve heart rate monitoring results based on individual and environmental conditions in which a device operates. Let's say two individuals wearing identical smartwatches from the same manufacturer and version while participating in a physical activity like running. The heart rate monitoring effectiveness on a wrist-based device depends on the skin contact quality, skin tone and tattoos, or the fit of the device. For instance, skin tone and the presence of tattoos can affect the optical sensors' ability to measure heart rate accurately. Darker skin tones and tattoos may require additional data or produce less accurate readings. More specifically, the environment plays a significant role in the accuracy of data collected by the smartwatch or bracelet. Nevertheless, manufacturers strive for consistency and accuracy while individual variations and environmental factors can lead to different results between two devices of the same version and manufacturer. 

Regarding the above-mentioned example, Figure~\ref{fig:fl_com} presents a scenario of FL within CommunityAI where communities are created from devices with the same data structures, e.g., aiming to improve heart rate monitoring results via anomaly detection (i.e., learning tasks). Embedded sensors or devices generate data for learning tasks for an FL community (i.e., health monitoring community). In the presented scenario, an FL population corresponds to all tasks with the same configurations (i.e., \textit{M2.1} and \textit{M2.2} belong to \textit{FL population 2}, and since they belong to the same \textit {Community C2}). As we witnessed in the example, environmental factors can lead to different results between two devices of the same version and manufacturer. This can cause a negative knowledge transfer by the model updates and decrease the overall model performance. Therefore,  FL cohorts are considered subsets of an FL population (as explained in \cite{hiessl2020industrial, hiessl2022cohort}). This enables knowledge sharing exclusively within, e.g., \textit{FL cohort 2}, including \textit{M2.2} models. Lastly, sharing updates between FL clients dealing with the same data characteristics and environmental conditions enhances the accuracy of their individual models.

\subsection{Architecture and Processes} 
\label{fl_arch}

\begin{figure*}[!t]
	\begin{center}
		\includegraphics[width=0.9\linewidth]{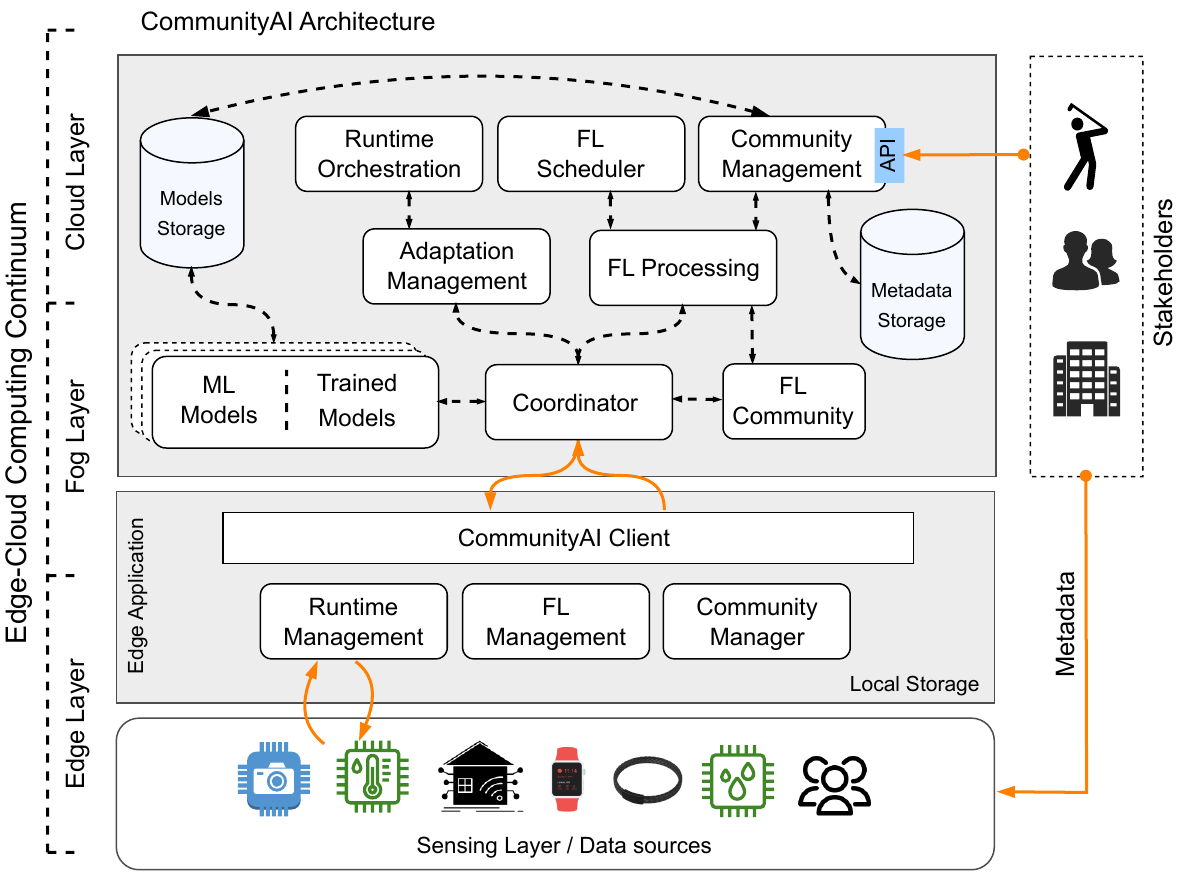}
		\caption{The conceptual architecture and workflow of CommunityAI.}
		\label{fig:workflow}
	\end{center} 
\end{figure*}

As illustrated in Figure~\ref{fig:workflow}, Edge Computing serves as a critical architectural intermediary layer between the cloud and end-users. When it comes to a platform for enabling training and deploying a ML model, it can harness the decentralized characteristics of such infrastructures. Furthermore, such infrastructures can support handling the massive data transfer, which may overcome latency issues. Therefore, the CommunityAI is designed as a three-tier architecture where software components can be deployed within the Edge-Cloud infrastructure. 

The CommunityAI architecture is composed of \texttt{(i)} server-side software components (i.e., which can be deployed and run in cloud and fog layers)  and \texttt{(ii)} client-side software components (i.e., which can be deployed and run in client devices such as smartphones or edge/fog devices in proximity to an end-user). From the infrastructure perspective, the cloud environment offers "unlimited" resources and advanced capabilities for orchestration, model execution, and managing the resources required. The fog layer comprises a set of stationary and powerful devices which can be physical or virtual, featuring various hardware configurations such as CPU, GPU, storage, and more. This layer has a supportive role in storing, training, and keeping often-used models near clients. The edge layer consists of low-powered devices that can make requests to the system (i.e., users ask for recommendations) or act as participating client device that contributes their local data and computational resources to the FL training process. These devices typically have limited resources, relying on CPUs and batteries for power, but often, they possess sufficient hardware capabilities to execute model training operations. Within the edge layer, we have embedded computing devices that generate various data information about the environment or data for specific purposes. As illustrated in Figure \ref{fig:workflow}, the server-side software components are \texttt{(i)} run-time Orchestration,\texttt{(ii)} FL Scheduler, \texttt{(iii)} Community Management, \texttt{(iii)} Adaptation Management, \texttt{(iv)} FL Processing, \texttt{(v)} ML Base and Trained Models, \texttt{(vi)} Coordinator, and \texttt{(vii)} FL Community. First, we describe software components that deal with architecture considerations such as run-time aspects, deployment, monitoring, adaptation, and resource management. Secondly, we explain the software components related to the FL process within the CommunityAI framework. Lastly, we describe the client-side software components.

\subsubsection{Software Architecture Considerations} The run-time orchestration component determines the optimal software component placement to ensure reliable and low-latency service delivery to end-users. Software components (e.g., adaptation management software sub-components) can be deployed geographically closer to end-users, such as fog devices for better service delivery. Furthermore, this component monitors the overall system and deployment of software components in the Edge-Cloud infrastructure. In addition, the run-time orchestration component determines where to deploy often-used ML models in proximity to clients so that the communication cost is optimized. The community management component provides an API that allows various stakeholders to create communities. The stakeholders are trustable end-users who can create various communities, define their metadata (e.g., collaboration criteria, preferences, etc.), and submit tasks using the provided API. Both device/community metadata files and ML base models are stored within their respective databases. The adaptation management component enhances flexibility and responsiveness to changing conditions and user needs. More specifically, it automatically allocates computational resources to scale up or down based on the user number, data volume, and processing demands. In other words, it provides several functions, such as resource allocation and scaling (i.e., elasticity \cite{murturi2022decent}), identifying and responding to security breaches or unauthorized access attempts, and ensuring the system remains efficient and responsive. Lastly, the coordinator component is responsible for several tasks within the CommunityAI framework. First, this component enables communication with other devices and users, registering them to the system, and getting metadata files from these resources. Secondly, the coordinator component shares a requested ML-base model with the registered clients (i.e., the sharing process is explained in Section \ref{fl_process}). Essentially, the coordinator component provides a communication channel for transmitting the model weights between the central server and the participating clients in a secure way. 

\subsubsection{FL Components and Process}
\label{fl_process}
FL components within the server-side of the CommunityAI are \texttt{(i)} FL Scheduler, \texttt{(ii)} FL Processing, and \texttt{(iii)} FL Community. A detailed lifecycle of a trained model in an FL system is given in \cite{li2023federated}. Initially, a typical FL workflow is typically driven by a model engineer (i.e., a stakeholder) who defines the problem to be solved and develops a suitable model architecture for FL. Clients generate the data for model training as well as submit FL tasks with metadata (i.e., including details such as targeted devices - device or resource a task is meant to operate on). The FL Scheduler component is responsible for mapping the task into the corresponding FL population and then providing scheduling instructions to initiate FL task execution with that FL population. FL Processing component converts the FL task to an FL plan. After translating the FL task into a plan, the FL Processing generates a global ML model suitable for the task. This model is typically created based on the specifics of the FL plan. Afterward, with the global ML model in place, the FL Processing initiates the FL process for a given FL cohort by establishing connections with all the FL clients within that cohort who have FL tasks to perform. FL Community component is responsible for maintaining such information. On the other side, similar to the FL Processing that operates on the server-side, there is a corresponding component on the client-side (see Section \ref{sec: client}). The resulting metrics from these client-side operations are provided to the FL Community component and used to update and manage cohorts.

\subsubsection{CommunityAI Client}
\label{sec: client}
The client-side software components are composed of four software components \texttt{(i)} CommunityAI Client, \texttt{(ii)} Runtime Management, \texttt{(iii)} FL Management, and \texttt{(iv)} Community Manager. Runtime Management contains a set of crucial responsibilities essential for maintaining the efficient operation of a device or system. More specifically, it continuously monitors internal hardware metrics, discovers and monitors nearby resources such as sensors, etc. Moreover, it gets metadata files from other nearby devices and shares them with the coordinator via the CommunityAI Client component.  The FL Management component provides a set of software tools that enable the execution of instructions provided by the FL Processing. These instructions typically pertain to tasks such as training or evaluating machine learning models on local edge devices. 

Essentially, the FL Management component plays a role similar to the FL Processing but is responsible for tasks on individual client devices.  When client devices execute evaluation plans specified in the FL plan, they generate performance metrics or measurements. The Community Manager shares these metrics with the FL processing component via the coordinator. The FL Processing shares these metrics as well as with the FL Community (i.e., server-side components). Furthermore, training can also be delegated to nearby edge devices through CommunityAI Client. Outsourcing or offloading a training process to these trusted edge devices with available resources becomes a practical solution. It ensures that additional training or computation can continue without overburdening a single-edge device. Lastly, note that a CommunityAI Client can be a stakeholder who initiates a community (i.e., submits tasks, models, etc.), or a user with data that joins a community and participates in the training process and subsequently uses such models on their machines.




\section{Research Challenges and Future Directions}
\label{sec:futurechallenges}
CommunityAI aims to facilitate the seamless sharing and knowledge transfer within FL communities. This can involve information dissemination, expertise, and best practices, helping community participants learn from each other. In other words, CommunityAI aims to harness AI power by enabling collaboration to enhance the way communities function, enabling them to thrive, adapt, and address complex challenges. Nevertheless, forming FL communities using FL techniques and orchestrating architecture components in dynamic environments introduces several challenges. We identify {four} main research directions that must be further investigated in the future:

\subsubsection{Metadata Model and Sharing Protocol} 
Metadata refers to information about communities, devices, data, or clients that is not the raw data itself but provides context or criteria for collaboration. In the FL context, this could include information about the types of data a client has, its expertise, its willingness to collaborate, or any specific criteria it requires for participating in FL tasks. Furthermore, such metadata information should be represented in a structured way, such as the format, attributes, and semantics of the metadata that FL clients can share with each other. Such a model would allow stakeholders to describe themselves and their collaboration preferences standardized and securely. Another important aspect is sharing protocol. Essentially, sharing protocol is a set of rules and procedures dictating how FL clients securely and selectively exchange metadata. Hence, it is important to explore methods, communication channels, and security approaches that might be involved in the exchange process.

\subsubsection{Advanced Community Filtering} 
The CommunityAI framework organizes participants into communities based on shared interests, expertise, data similarities, or characteristics. However, there is a need for more advanced and adaptive methods to identify and establish such communities. Moreover, it is crucial to acknowledge that participants within communities may change their characteristics over time. These changes can result from evolving interests, skill development, shifting device usage, etc. Therefore, it is important to explore advanced community detection algorithms that consider multiple characteristics as well as adapt to evolving participant characteristics. Additionally, verifying the accuracy and effectiveness of community identification processes remains an important issue that must be addressed.

\subsubsection{Predicting Negative Knowledge Transfer}
Negative knowledge transfer may occur when information from one participant adversely affects the performance of another participant's model. This can result in decreased model quality. By predicting such instances, FL communities can proactively take steps to mitigate these effects, ensuring high model quality. Therefore, it is worth exploring methods that detect potential negative knowledge transfer early, preventing unnecessary training iterations, data exchanges, and resource usage. This enhances the efficiency of the FL process, particularly on resource-constrained devices. However, developing accurate methods for predicting negative knowledge transfer is a significant challenge. 

\subsubsection{Privacy-preserving CommunityAI}
FL effectively tackles privacy concerns through local data storage; however, supplementary privacy-preserving methods are needed to guarantee the non-disclosure of sensitive information during the model aggregation phase. Hence, it is important to explore methods for designing resilient and privacy-conscious models that can exhibit strong performance across various domains while safeguarding the sensitive data of individual clients or domains. Furthermore, trust establishment within a community represents another significant challenge that requires further exploration in developing advanced methods.

\section{Conclusion}
\label{sec:conlusion}
CommunityAI is an FL-based framework designed to group participants into communities based on shared interests, expertise, or data characteristics, fostering collaboration in various domains like health, education, industrial automation, and other domains. Participants within the framework range from computing devices and sensors to humans working together in these communities to train and improve machine learning models collaboratively. Notably, the FL approach and collaborative process prioritizes data and participant privacy by ensuring that data is shared and used only within their respective groups or communities. Within this paper, we presented vision aspects and represented an initial step toward introducing the CommunityAI framework. In our forthcoming research, our goal is to deliver a comprehensive technical framework that encompasses both technical and architectural elements. Additionally, the research challenges we have identified require thorough investigation in the future.
 
\section*{Acknowledgment}
Research has partially received funding from grant agreement No. 101079214 (AIoTwin) and by EU Horizon Framework grant agreement 101070186 (TEADAL).

\bibliographystyle{IEEEtran}
\bibliography{ref.bib}

\end{document}